\documentclass{article}

\usepackage{PRIMEarxiv}

\usepackage{amssymb}
\usepackage{amsmath,amsfonts}
\usepackage[utf8]{inputenc} 
\usepackage[T1]{fontenc}    
\usepackage{hyperref}       
\usepackage{stfloats}
\usepackage{url}            
\usepackage{bbm}
\usepackage{booktabs}       
\usepackage{amsfonts}       
\usepackage{nicefrac}       
\usepackage{microtype}      
\usepackage{lipsum}
\usepackage{fancyhdr}       
\usepackage[caption=false,font=normalsize,labelfont=sf,textfont=sf]{subfig}
\usepackage{graphicx}       
\usepackage{booktabs}
\usepackage{epsfig}
\usepackage{multirow} 
\DeclareMathOperator*{\sign}{sign}
\usepackage{algorithm,tabularx} 
\usepackage{algorithmic}

\graphicspath{{media/}}     

\title{FedCCL: Federated Dual-Clustered Feature Contrast Under Domain Heterogeneity
}

\author{
Yu Qiao\\
Kyung Hee University \\
Republic of Korea \\
\texttt{qiaoyu@khu.ac.kr} \\
\and
Huy Q. Le \\
Kyung Hee University \\
Republic of Korea \\
\texttt{quanghuy69@khu.ac.kr} \\
\and
Mengchun Zhang \\
Kyung Hee University \\
Republic of Korea \\
{\texttt{mengchun0527@gmail.com}} \\
\and
Apurba Adhikary \\
Kyung Hee University \\
Republic of Korea \\
\texttt{apurba@khu.ac.kr} \\
\and
Chaoning Zhang \\
Kyung Hee University \\
Republic of Korea \\
{\small\texttt{chaoningzhang1990@gmail.com}} \\
\and
Choong Seon Hong\thanks{Corresponding author: Choong Seon Hong}\\
Kyung Hee University \\
Republic of Korea \\
\texttt{cshong@khu.ac.kr} \\
}

\begin{document}
\maketitle

\begin{abstract}
Federated learning (FL) facilitates a privacy-preserving neural network training paradigm through collaboration between edge clients and a central server. One significant challenge is that the distributed data is not independently and identically distributed (non-IID), typically including both intra-domain and inter-domain heterogeneity. However, recent research is limited to simply using averaged signals as a form of regularization and only focusing on one aspect of these non-IID challenges. Given these limitations, this paper clarifies these two non-IID challenges and attempts to introduce cluster representation to address them from both local and global perspectives. Specifically, we propose a dual-clustered feature contrast-based FL framework with dual focuses. First, we employ clustering on the local representations of each client, aiming to capture intra-class information based on these local clusters at a high level of granularity. Then, we facilitate cross-client knowledge sharing by pulling the local representation closer to clusters shared by clients with similar semantics while pushing them away from clusters with dissimilar semantics. Second, since the sizes of local clusters belonging to the same class may differ for each client, we further utilize clustering on the global side and conduct averaging to create a consistent global signal for guiding each local training in a contrastive manner. Experimental results on multiple datasets demonstrate that our proposal achieves comparable or superior performance gain under intra-domain and inter-domain heterogeneity.
\end{abstract}


\section{Introduction}
\label{sec:intro}
Advancements in artificial intelligence have revolutionized the way and efficiency of data processing, including centralized neural network training. This training method commonly requires aggregating large amounts of available data, which is usually generated by various real-world sources such as financial institutions, enterprises, and numerous Internet of Things devices distributed at edge nodes~\cite{khan2021federated,nguyen2021federated}. However, privacy and security concerns arise with this training method because the data involved are inherently privacy-sensitive. Additionally, transmitting raw data for centralized training can lead to communication overhead issues~\cite{kairouz2021advances,tun2023contrastive}. In response to these challenges, federated learning (FL) stands as a promising distributed machine learning technology that can facilitate distributed training of neural network models through the collaboration of edge devices and a central server without compromising the privacy of sensitive data~\cite{mcmahan2017communication}. Nevertheless, FL still confronts a significant challenge, i.e., the data distribution among clients is not independently and identically distributed (non-IID)~\cite{zhang2022federated}. Within this challenge, each client optimizes the local model based on its own local private data, which may not be consistent with the updated direction of the global model. Consequently, the optimized local models tend to converge slowly and exhibit a tendency to overfit their own data, further compromising their generalization ability~\cite{tan2022federated}.

\begin{figure}[t]
\centering
\includegraphics[width=0.60\textwidth]{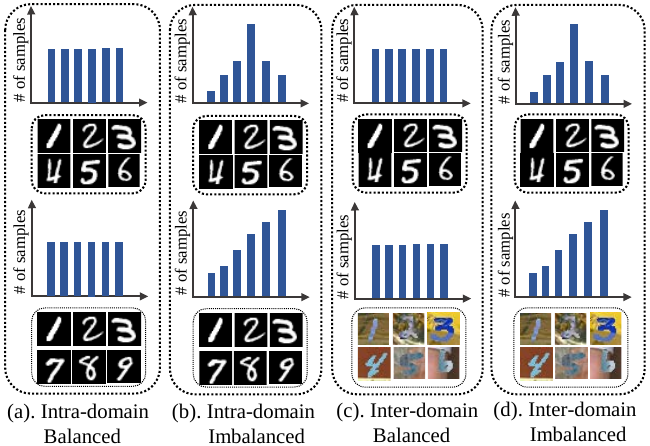}
\caption{Illustration of four challenges in FL. Figure (a) illustrates \textit{balanced intra-domain heterogeneity}, where samples are drawn from the same domain but exhibit label shifts with equal quantities. Figure (b) shows \textit{imbalanced intra-domain heterogeneity}, where samples are drawn from the same domain but with varying quantity and label shifts. Figure (c) denotes \textit{balanced inter-domain heterogeneity}, where samples are drawn from different domains but with the same sample quantities. Figure (d) represents \textit{imbalanced inter-domain heterogeneity}, where samples are drawn from different domains with varying sample quantities.}
\label{fig:challenges}
\end{figure}

To address the challenges mentioned above, numerous studies~\cite{zhu2021federated,mu2023fedproc,tan2022fedproto,gao2022feddc,karimireddy2020scaffold} concentrate on addressing label non-IID (a.k.a balanced intra-domain heterogeneity, see Figure~\ref{fig:challenges} (a)) by introducing a regularization term into the local objective for each client. This addition serves to guide the training process at each local level and enhance the model's generalization capability to unseen data. However, despite its efficacy, such approaches often focus solely on addressing this particular aspect of the challenges, while neglecting another significant challenge: imbalanced intra-domain heterogeneity: imbalanced intra-domain heterogeneity (see Figure~\ref{fig:challenges} (b)) and balanced / imbalanced inter-domain heterogeneity (see Figure~\ref{fig:challenges} (c), (d)). For example, in the presence of imbalanced intra-domain heterogeneity challenge (Figure~\ref{fig:challenges} (b)), local model optimization may exhibit a bias towards dominant labels due to their larger sample sizes. Similarly, when addressing inter-domain heterogeneity challenges (Figure~\ref{fig:challenges} (c), (d)), optimization may exhibit biases towards the dominant domain. To address these challenges, a common practice is to regularize the local model using a global signal. This signal is typically obtained by averaging the representations from each client to derive the local signal and then further averaging all these local signals belonging to the same semantics. However, one limitation of this approach can be noted that it primarily relies on the average value to represent intra-domain information as well as global signals, thereby overlooking the fine-grained nature of intra-domain data and the potential inter-domain dominance among clients.

Considering the aforementioned limitation, we rethink the generation of local and global signals~\cite{huang2023rethinking,mu2023fedproc,li2020prototypical}, which are usually defined as the mean value of features belonging to the same class space and are represented as a vector~\cite{tan2022fedproto}. In particular, the local signal denotes the average feature values within the same semantic context, while the global signal represents the average of the local signal that belongs to the same class space.  Nevertheless, simply averaging features with identical semantics to get local signals would result in the model's failure to capture critical distinctions within each class, leading to imprecise feature representations and a decline in generalization performance~\cite{li2021adaptive}. Moreover, the global signals may introduce biases towards dominant local signals, primarily due to the imbalance in the generated local signals, potentially leading to unrepresentative global signals. Motivated by these two issues, we first derive local signals through clustering features within the same class space. Therefore, each class is expected to be represented by fine-grained local signals, allowing the model to capture variations and characteristics within each class effectively. Second, considering that the size of generated local cluster signals from each client varies, directly averaging the local cluster signals may introduce bias towards dominant clients. Therefore, we hypothesize that conducting clustering on the local clustered signals from all clients can address this issue. This is because the resulting clustered global signals can represent the contribution from each client fairly~\cite{abbasi2021fair}, potentially ensuring a relatively more equal representation of each client's contribution. Moreover, we argue that these global cluster signals need to be averaged again to obtain a consistent signal. Given that global signals contain information from different clients, learning from global signals can effectively capture coarse-grained knowledge across diverse local models at a high level. Consequently, the model is expected to better handle similarities and differences among classes, thus enhancing generalization performance. We note that in contrast to raw feature vectors, local and global signals are only one-dimensional vectors, which are averaged several times. Therefore, they are more privacy-friendly, and it is difficult to recover the raw data from them~\cite{tan2022fedproto,huang2023rethinking}.

In this paper, we propose \textbf{Fe}derated \textbf{d}ual-\textbf{C}lustered feature \textbf{C}ontrastive \textbf{L}earning (\textbf{FedCCL} for short), which includes two key components. First, we introduce local clustered feature contrast, which utilizes local clustered signals to facilitate cross-client contrastive learning~\cite{he2020momentum,chen2020simple,wu2018unsupervised,tan2022federated_new,chen2021exploring}. It can encourage each representation to closely align with local signals belonging to the same semantics across different clients while simultaneously pushing it away from the clusters with different semantics. In other words, the local clustered signals not only enhance the model's ability to capture a high level of granularity within each class but also promote effective knowledge sharing among clients. Second, we construct a consistent global signal and propose global clustered feature contrast, which utilizes the unbiased global signal to provide guidance for each local training. To elaborate, on the global side, we cluster local clustered signals to derive global cluster signals and then average these global cluster signals to obtain their corresponding consistent global signals. Each local representation is encouraged to be close to its respective global signal within the same semantic class while simultaneously diverging from signals with different semantics. This approach is expected to help mitigate the risk of overfitting to local optima, thereby enhancing the generalization of local models. Our main contributions are summarized as follows:
\begin{itemize}
\item We first outline the non-IID challenges from both intra-domain and inter-domain perspectives. Subsequently, we propose a new federated framework that incorporates a dual-clustered feature contrast strategy aimed at learning well-generalizable local models. To the best of our knowledge, this is the first proposal of a dual-clustered feature contrast strategy in FL.
\item The proposed dual-clustered feature contrast FL framework consists of two new components. First, the local clustered feature contrast captures intra-class information for each client at a high granularity level. Second, the global clustered feature contrast guides each local training, enhancing the generalization of local models.
\item We experiment on various datasets to evaluate the effectiveness of our proposed method in addressing both intra-domain and inter-domain heterogeneity challenges. The results demonstrate that our approach achieves comparable or superior performance compared to several baselines. Moreover, we conduct a privacy protection evaluation in terms of adversarial attacks, revealing that our proposal is more robust than the robust FL baseline.
\end{itemize}

The subsequent sections of this paper are structured as follows: Section \ref{sec:relatedwork} presents related work. Section~\ref{sec:preliminary} provides the notation and preliminaries. Section~\ref{sec:methods} outlines the methods used. Section~\ref{sec:experiments} presents the experimental results. Finally, Section~\ref{sec:conclusion} draws conclusions.

\section{Related Work}
\label{sec:relatedwork}
In this section, we review the non-IID challenges in FL and discuss some FL-related methods in Section~\ref{subsec:FL_non_iid}. Following that, we briefly summarize progress in contrastive learning in Section~\ref{subsec:contrastive_learning}.
\subsection{Federated Learning}
\label{subsec:FL_non_iid}
FL is a solution designed to address privacy concerns in distributed machine learning, but it faces the challenge of data heterogeneity (also known as the non-IID issue). This challenge arises from variations in data distribution among participating clients and can potentially limit the effectiveness of FL. The pioneering work in FL, known as FedAvg~\cite{mcmahan2017communication}, introduced the training paradigm of FL. FedAvg trains a global model through the collaborative efforts of local clients, each of which possesses heterogeneous data. In line with this approach, various existing methods~\cite{wang2020tackling,t2020personalized,karimireddy2020scaffold,yao2019federated,li2021model,li2020federated} have been introduced to address this data heterogeneity.  FedProx~\cite{li2020federated}, MOON~\cite{li2021model}, and pFedME~\cite{t2020personalized} introduce an approach in which global model parameters can serve as a reference to adjust local training processes, effectively mitigating local update biases. In addition, FedNova~\cite{wang2020tackling} suggests that local clients should conduct different local epochs while updating their local models. SCAFFOLD~\cite{karimireddy2020scaffold} introduces two control variables that encapsulate the updated directional information from global and local models, thereby overcoming gradient inconsistency. FedMMD~\cite{yao2019federated} extracts a more generalized representation by minimizing the maximum mean difference loss through a two-stream model. CSAC~\cite{yuan2023collaborative} proposes a collaborative semantic aggregation and calibration framework to achieve domain generalization. 

Moreover, researchers have also been dedicated to addressing the non-IID issue in FL by adopting a representative learning perspective to mitigate biases in local models. One representative method is the utilization of prototypes for representation learning. The prototype concept was originally introduced by prototype networks~\cite{snell2017prototypical} in the domain of few-shot learning~\cite{wang2020generalizing}. The core idea behind prototype learning is to utilize a single prototype to represent each class, which is calculated by averaging the embedding vectors within the same class space. Notably, prototype learning has witnessed significant success in various domains, including medical image analysis~\cite{cheng2023prior}, semantical segmentation~\cite{dong2018few}, and natural language processing~\cite{wieting2015towards}. Very recently, the concept of prototypes has been extended to the domain of FL, with the consideration of privacy protection~\cite{tan2022fedproto,huang2023rethinking} and communication efficiency~\cite{qiao2023mp_arxiv}. As one notable seminal work, FedProto~\cite{tan2022fedproto} suggests reducing the data exchange between clients and the server only by communicating prototypes instead of the traditional method of exchanging gradients or model parameters. However, their method has not been validated in a broader and diverse heterogeneous environment, such as one characterized by the Dirichlet distribution~\cite{yurochkin2019bayesian}. However, the main limitation of the above methods is that they only focus on one aspect of non-IID challenges, while ignoring the inter-domain heterogeneity challenges~\cite{li2020adaptive,li2021adaptive}. Additionally, some of these methods require heavy communication of model parameters. In this paper, we address three practical and challenging scenarios where data heterogeneity among clients includes imbalanced intra-domain, balanced inter-domain, and imbalanced inter-domain challenges. Moreover, we utilize clustered features for communication, which are considerably more efficient in terms of size compared to model parameters~\cite{huang2023rethinking,dai2023tackling}.

\subsection{Contrastive Learning}
\label{subsec:contrastive_learning}
Contrastive learning is a paradigm of unsupervised learning that learns powerful representations by comparing positive sample pairs and negative sample pairs within the embedding space. Its primary objective is to maximize the similarity between positive sample pairs while minimizing the similarity between positive and negative samples. MoCo~\cite{he2020momentum} and SimCLR~\cite{chen2020simple} represent two notable approaches capable of achieving performance comparable to supervised learning. Specifically, MoCo employs the InfoNCE~\cite{oord2018representation} objective function, marking a significant milestone as it outperforms supervised methods in unsupervised learning for the first time, with its core strategy involving the implementation of a momentum update to build a more robust negative sample set. Besides, SimCLR introduces a data augmentation technique to increase the number of positive sample pairs by generating multiple distinct augmented views from the same image. Since then, contrastive learning and its variants have been applied to different domains~\cite{you2020graph,kuang2021video,spijkervet2021contrastive,saeed2021contrastive}. 

In addition, several works~\cite{tan2022federated,li2021model,mu2023fedproc,huang2023rethinking} have extended the application of contrastive learning to the FL domain. FedProc~\cite{mu2023fedproc} and FPL~\cite{huang2023rethinking} propose an approach where global prototypes on the server serve as references for guiding the training of clients during local updates. They employ a contrastive loss function to encourage each class to converge towards its respective global prototype while pushing it away from other global prototypes. However, these works are limited to addressing a single challenge in FL and only improve the generation of the global signal based on the average local representation of each client. We argue that this direct averaging of local representations may still introduce bias and further contribute to biases in the global signal. In contrast to previous work, we deviate from the common practice of using averaging for local signal generation and instead opt for clustering. Second, for the generation of global signals, we employ two clustering operations followed by averaging. Moreover, we introduce a dual-side clustered feature contrast strategy, comprising local clustered feature contrast and global clustered feature contrast. This strategy aims to furnish local models with rich knowledge and enhanced generalization capabilities to tackle intra-domain and inter-domain heterogeneity challenges.

\begin{figure*}[t]
\centering
\includegraphics[width=\textwidth]{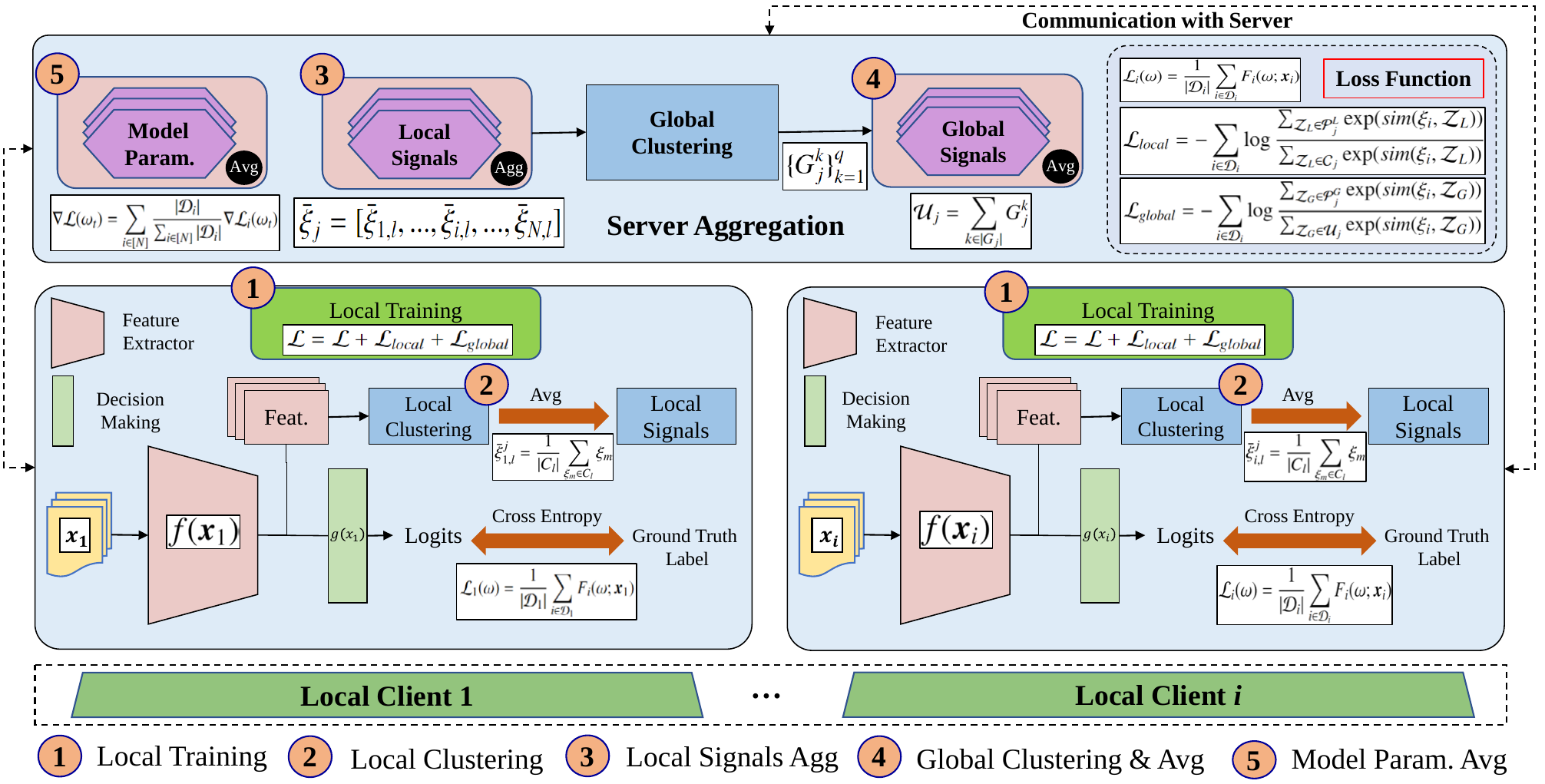}
\caption{Illustration of our proposed FedCCL framework. The proposed framework comprises five main steps: local training, local clustering, local signal aggregation, global signal aggregation and averaging, and global model parameter averaging. In this paper, we focus on steps 2 to 4.}
\label{fig:system_model}
\end{figure*}

\section{Notation and Preliminaries}
\label{sec:preliminary}
In this section, we first outline the challenges of non-IID from an intra-domain and inter-domain perspective in Section~\ref{subsec:non_iid_classification}, followed by outlining the basic setup of FL in Section~\ref{subsec:notation_fl}. 

\subsection{Domain Heterogeneity Challenges}
\label{subsec:non_iid_classification}
The design objective of FL is to collaboratively train a shared model using decentralized data while ensuring data privacy protection, and it can be applied to distributed edge clients. However, the presence of data heterogeneity among these clients usually results in models trained through federated training becoming biased towards dominant data sources. Considering a scenario with $N$ clients, each indexed by $i$, with their respective dataset distributions denoted as $\mathcal{D}_i$ for pairs of samples {($\boldsymbol x_i$, $y_i$)}, where $\boldsymbol x_i$ represents the sample, and $y_i$ represents the label associated with the sample. The label distribution on different clients can be represented as $P_i(y)$, and feature distribution on different clients can be denoted as $P_i(\boldsymbol{x})$. In addition, the size of each client's dataset can be represented as $|\mathcal{D}_i|$.

Before introducing FL-related technologies, we first outline the non-IID challenges in FL from a domain perspective. This approach is adopted to provide a relatively more straightforward understanding of the diverse challenges posed by non-IID data, particularly highlighting the variations within and between different domains. Specifically, we categorize non-IID challenges into four main categories: (1) \textit{balanced Intra-domain heterogeneity}, where distributed data is sampled from the same domain but exhibits label shift with equal quantities, (2) \textit{imbalanced Intra-domain heterogeneity}, where data is sampled from the same domain but with different label shift and quantities, (3) \textit{balanced Inter-domain heterogeneity}, where data is sourced from different domains but has the same label quantities, and (4) \textit{imbalanced Inter-domain heterogeneity}, where data is sampled from different domains with varying sample quantities. Figure~\ref{fig:challenges} illustrates these four challenges for better visualization, and these challenges can be summarized as follows: 
\begin{itemize}
\item \textbf{Balanced intra-domain heterogeneity}. Data sampled from the same domain with label shift but equal quantities. the feature distributions $P_i(\boldsymbol{x})$ and dataset size $|\mathcal{D}|$ on different clients are similar, while the label distributions $P_i(y)$ differ. Specifically, $P_i(\boldsymbol{x}) = P_j(\boldsymbol{x})$ and $|\mathcal{D}_i| = |\mathcal{D}_j|$, but $P_i(y) \neq P_j(y)$ for clients $i$ and $j$.
\item \textbf{Imbalanced intra-domain heterogeneity}. Data from the same domain with different label shifts and quantities. Mathematically, both dataset size $|\mathcal{D}|$ and label distributions $P_i(y)$ on different clients are dissimilar, while feature distributions $P_i(\boldsymbol{x})$ are the same. Specifically, $P_i(\boldsymbol{x}) = P_j(\boldsymbol{x})$, $|\mathcal{D}_i| \neq |\mathcal{D}_j|$, and $P_i(y) \neq P_j(y)$ for clients $i$ and $j$.
\item \textbf{Balanced inter-domain heterogeneity}. Data sourced from different domains with the same label quantities. Mathematically, the feature distributions $P_i(\boldsymbol{x})$ on different clients are dissimilar, while dataset size $|\mathcal{D}|$, and label distributions $P_i(y)$ are the same. Specifically, $|\mathcal{D}_i| = |\mathcal{D}_j|$, $P_i(y) = P_j(y)$ but $P_i(\boldsymbol{x}) \neq P_j(\boldsymbol{x})$ for clients $i$ and $j$.
\item  \textbf{Imbalanced inter-domain heterogeneity}. Data sampled from different domains with varying sample quantities. Mathematically, dataset size $|\mathcal{D}|$, feature distributions $P_i(\boldsymbol{x})$, and label distributions $P_i(y)$ on different clients are dissimilar. Specifically, $|\mathcal{D}_i| \neq |\mathcal{D}_j|$, $P_i(\boldsymbol{x}) \neq P_j(\boldsymbol{x})$, and $P_i(y) \neq P_j(y)$ for clients $i$ and $j$.
\end{itemize}

In this paper, we attempt to address the more challenging imbalanced intra-domain heterogeneity (Figure~\ref{fig:challenges} (b)), as well as the inter-domain heterogeneity challenges (Figure~\ref{fig:challenges} (c) and (d)), which are usually overlooked by the FL community but are considered both practical and challenging~\cite{li2021fedbn,li2022federated}.

\subsection{Federated Learning}
\label{subsec:notation_fl}
The FL framework aims to collaboratively train a well-performing global model by leveraging the distributed data from clients while ensuring data privacy protection at the local level. We denote the local model for each client as $F_i(\omega; \boldsymbol{x}_i)$, where $\omega$ represents the model parameters. The loss function (such as cross-entropy loss) for each client can then be defined as follows:
\begin{equation} \label{one_hot}
   {\mathcal{L}_i} (\omega; \boldsymbol{x}_i) = -\sum_{j=1}^{\mathbb{C}}\mathbbm{1}_{y = j}\log p_j(F(\omega; \boldsymbol {x}_i); y_i),
\end{equation}
where $\mathbbm{1}(\cdot)$ represents the indicator function, $p_j(\cdot)$ denotes the model predictions after the softmax operation, and $\mathbb{C}$ is the set of classes, containing a total of $C$ classes. Then, the local training process for each client $\mathcal{L}_i$ can be defined as:
\begin{equation} \label{local_loss}
   {\mathcal{L}_i (\omega)} = \frac {1}{|\mathcal{D}_i|} \sum_{i \in \mathcal{D}_i} F_i(\omega; \boldsymbol{x}_i).
\end{equation}

During the FL training process, each client participates in training and performs local optimization to update its local weights, and $\omega_{t}$ denotes the updated parameters of the global model from the previous round.
\begin{equation} \label{local_sgd}
 \omega_{t}^i  =  \omega_{t}^i - \eta \nabla \mathcal{L}_i(\omega_{t-1}),
\end{equation}
where $\eta$ denotes the learning rate, and $\nabla \mathcal{L}_i(\omega_{t-1})$ is the local gradient for each client. 

After each client finishes updating its local model parameters and sends them back to the server for aggregation, the global model parameters can then be updated as follows:
\begin{equation} \label{global_weight}
\nabla \mathcal{L}(\omega_{t}) = \sum_{i \in [N]}\frac{|\mathcal{D}_i|}{\sum_{i \in [N]}|\mathcal{D}_i|} \nabla \mathcal{L}_i(\omega_{t}).
\end{equation}

Overall, the training optimization process involves minimizing the local loss across distributed clients. This process can be summarized as follows:
\begin{equation} \label{global_loss}
   \mathop {\arg\min}_{\omega_t} {\mathcal{L}(\omega_t)},
\end{equation}
where $\omega_t$ represents the aggregated model parameters in global round $t$, which are subsequently sent back to local clients for the next iteration until convergence.

\section{Methodology}
\label{sec:methods}
In this section, we first introduce local clustered feature generation in Section~\ref{subsec:local_signals}, followed by global clustered feature generation in Section~\ref{subsec:global_signals}. Finally, we present the proposed local updating process in Section~\ref{sub:proposed_local_updates}.
\subsection{Local Clustered Feature Representation}
\label{subsec:local_signals}
Deep neural networks (DNNs) generally comprise two main components: the feature extraction layer $\boldsymbol{f}$ and the classification layer $\boldsymbol g$. The former primarily encodes inputs into embeddings, while the latter makes final predictions based on the learned features. For a given client $i$, the embedding feature of its $j$-th class instance can be calculated as follows: 
\begin{equation} \label{feature_extrac}
\xi_{i,j} = \{\boldsymbol f(\boldsymbol {x}_i) \,\,|\,\, \boldsymbol {x}_i \in \mathcal D_{i,j} \},
\end{equation}
where $\mathcal D_{i,j}$ is a subset of $\mathcal{D}_i$ belonging to the $j$-th class, and $\xi_{i,j}$ represents the embedding set of the $j$-th class. Previous methods often rely on directly averaging within the embedding space to obtain the local signal for each class, as shown below:
\begin{equation} \label{local_prototype}
{h}_{i,j} = \frac{1}{|\mathcal D_{i,j}|}\sum_{\boldsymbol {x}_i \in \mathcal D_{i,j}} \xi_{i,j}.
\end{equation}

However, given the diversity inherent in the datasets used to train DNNs, even samples belonging to the same class may exhibit varied distributions~\cite{em2017incorporating,zeng2020elm}. For instance, a set of samples belonging to the category of cars may include different brands, models, colors, and appearances. In such scenarios, relying solely on averaged local signals (computed directly from embeddings of the same class) may not adequately represent the data distribution. Therefore, we attempt to enhance the representation by leveraging hierarchical clustering~\cite{ran2023comprehensive} techniques. This is because hierarchical clustering has the capability to capture the inherent structure within the data, generating a clustering tree that offers meaningful interpretations of data at various levels of granularity~\cite{ran2023comprehensive,finch_clustering}.

Inspired by the success of hierarchical clustering adopted in various domains~\cite{liu2023finch,zhuang2021towards,huang2023rethinking,meng2023mhccl,sarfraz2019efficient}, we utilize the parameter-free clustering method FINCH~\cite{finch_clustering} which belongs to the family of hierarchical clustering and can automatically determine the number of clusters. Specifically, when given the input $\xi_{i,j}$ belonging to the $j$-th class for client $i$, we initially derive its adjacency link matrix using cosine similarity metrics, as illustrated by the following equation:
\begin{equation} \label{adjacency_matrix}
A_i^j(\xi_m,\xi_n) =
\begin{cases}
    1, & \text{if } n = k_{\xi_m}^1 \text{ or } k_{\xi_n}^1 = m \text{ or } k_{\xi_m}^1 = k_{\xi_n}^1 \\
    0, & \text{otherwise}
\end{cases},
\end{equation}
where $A_i^j(\xi_m,\xi_n)$ denotes the adjacency link matrix of client $i$ belongings to class $j$, and $k_{\xi_m}^1$ represents the first neighbor of input embedding feature $\xi_m$ which is one element of 
$\xi_{i,j}$.

After calculating the first neighbor for each instance in $\xi_{i,j}$ using Eq.~\ref{adjacency_matrix}, we obtain the initial partition containing multiple clusters. Subsequently, to merge these clusters recursively, we first adopt the strategy where we compute the mean vectors of the data samples within each cluster, which can be represented as follows:
\begin{equation} \label{average_first_partition}
P_k = \frac{1}{|C_k|} \sum_{\xi_m \in C_k} \xi_m,
\end{equation}
where $P_k$ represents the average vector of the $k$-th cluster, $C_k$ represents the set of data samples in the $k$-th cluster, and $\xi_m$ represents the embedded feature of the input sample.

The calculated mean vectors in Eq.~\ref{average_first_partition} are then utilized to determine the first neighbor cluster still based on the adjacency link matrix during cluster merging at each recursive step. This process continues until the current number of clusters no longer changes, or there is only one cluster left.  Finally, within each final cluster, we conduct averaging again. The resulting averaged value can be utilized to represent the local signal, as follows:
\begin{equation} \label{local_signal}
\bar{\xi}_{i, l}^j = \frac{1}{|C_l|} \sum_{\xi_m \in C_l} \xi_m,
\end{equation}
where $\bar{\xi}_{i, l}^j$ denotes the averaged value within the final cluster $l$ belongings to $i$-th client of $j$ class, and $C_l$ represents the set of data samples in the $l$-th final cluster. Note that averaged values are also commonly referred to as prototypes in various domains~\cite{ge2024beyond,du2023prototype,tan2022fedproto}. However, since our method of obtaining local signals differs from simply averaging values, we opt to utilize local signals to represent the averaged values in our approach.

Moreover, to enable a complete view of the data groupings discovered between clients, each client needs to send their locally clustered signal to the server for aggregation. Here, the aggregation is defined as follows:
\begin{align} \label{agg_clustered_local_signal}
\bar{\xi}_j &= [\bar{\xi}_{1, l},..., \bar{\xi}_{i, l},..., \bar{\xi}_{N, l}],
\end{align}
where $\bar{\xi}_j$ is the collected fine-grained view of data from all participating clients belonging to $j$-th class. These aggregations, along with the model parameters, are subsequently sent back to the local side for local training.

\subsection{Global Clustered Feature Representation}
\label{subsec:global_signals}
To mitigate biases between local models and the global model, a common practice is to derive a global signal and utilize it to guide each local training process. The global signal is usually derived by averaging the local clustered signals in Eq.~\ref{agg_clustered_local_signal} to obtain the corresponding global signals, as shown below:
\begin{equation} \label{avg_local_signals}
\mathcal H_{j} = \frac{1}{|\mathcal N_{j}|}\sum_{j \in \mathcal N_{j}} \bar{\xi}_j,
\end{equation}
where $\mathcal N_{j}$ denotes the sample belonging to class $j$, and $|\mathcal N_{j}|$ represents the size of $\mathcal N_{j}$. However, considering the potential inconsistency in the sizes of local clustered signals, we argue that directly averaging these signals to obtain the global signal may still introduce bias towards dominant groups. This bias arises due to the unequal contributions of different clusters to the global signal, which can skew the representation of the overall data distribution. 

To address this concern, we suggest performing another round of clustering on the locally clustered signals. By iteratively clustering the local signals, it can be expected that the global cluster centers obtained represent contributions from different clients more uniformly. Similar to the local clustering process, we define the global clustering process.
Specifically, when given the input $\bar{\xi}_{j}$ belonging to the $j$-th class, we derive its adjacency link matrix using cosine similarity metrics, as illustrated by the following equation:
\begin{equation} \label{global_adjacency_matrix}
A_j(\bar{\xi}_m,\bar{\xi}_n) =
\begin{cases}
    1, & \text{if } n = k_{\bar{\xi}_m}^1 \text{ or } k_{\bar{\xi}_n}^1 = m \text{ or } k_{\bar{\xi}_m}^1 = k_{\bar{\xi}_n}^1 \\
    0, & \text{otherwise}
\end{cases},
\end{equation}
where $A_j(\bar{\xi}_m,\bar{\xi}_n)$ denotes the adjacency link matrix of class $j$, and $k_{\bar{\xi}_m}^1$ represents the first neighbor of input embedding feature $\bar{\xi}_m$ which is one element of $\bar{\xi}_{i,j}$.

Similar to the process of generating local clusters, after obtaining the first neighbor for each instance in $\bar{\xi}_{i,j}$ using Eq.~\ref{global_adjacency_matrix}, we recursively merge these clusters by comparing the mean vectors within each partition. We repeat this process until the current number of clusters no longer changes, or there is only one cluster left. To simplify, the global clustering process is represented as follows:
\begin{equation} \label{cluster_global_signals}
[\bar{\xi}_{1, l},..., \bar{\xi}_{i, l},..., \bar{\xi}_{N, l}] \xrightarrow{Cluster} \{{G}_j^k\}_{k=1}^{q},
\end{equation}
where ${G}_j^k$ represents $q$ global clustered representative signals of $j$ class. However, the sizes of global clustered signals may still vary across different classes, which may not provide a consistent global signal for each local model training. Therefore, the global clustered signals are further averaged to obtain a single consistent global signal for each class, ensuring consistent guidance for each local training. Formally,
\begin{equation} \label{avg_cluster_global_signals}
\mathcal{U}_j = \sum_{k \in |{G}_j|} {G}_j^k.
\end{equation}
where $\mathcal{U}_j$ is an averaged global clustered signal belonging to $j$-th class, and $|{G}_j|$ denotes the size of ${G}_j$.

\begin{algorithm}[t]
    \caption{FedCCL}
    \label{alg:FedCCL}
    \begin{algorithmic}
        \REQUIRE Global communication rounds $T$, dataset $\mathcal{D}_i$ of each client, number of local clients $N$, local epochs $E$.
        \ENSURE Local model $\omega_i$ for each client.
    \end{algorithmic}
    \textit{Server executes:}
    \begin{algorithmic}[1]
    \FOR{$t = 1, 2, \ldots, T$}
        \FOR{$i = 0, 1, \ldots, N$ \textbf{in parallel}}
            \STATE {$\omega_i^t, \bar{\xi}_{i} \gets \textbf{LocalUpdate}(\omega_i^t, \mathcal{U})$}
        \ENDFOR
        \STATE $\bar{\xi}_j \gets [\bar{\xi}_{1, l}, ..., \bar{\xi}_{i, l}, ..., \bar{\xi}_{N, l}]$ by Eq.~\ref{agg_clustered_local_signal}
        \STATE {Global clustering based on Eqs.~\ref{global_adjacency_matrix} and~\ref{cluster_global_signals}}
        \STATE {Obtain global signals by Eq.~\ref{avg_cluster_global_signals}}
        \STATE {$\omega^{t+1} \gets \sum_{i=1}^{N} \frac{|\mathcal{D}_i|}{|\mathcal{D}|}\omega_i^t$}
    \ENDFOR
    \end{algorithmic}
    \textit{Local clients execute:}
    \begin{algorithmic}[1]
    \STATE \textbf{LocalUpdate($\omega_i^t$, $\mathcal{U}$):}
    \FOR{each local epoch}
        \FOR{each batch ($\boldsymbol{x}_i$; $y_i$) of $\mathcal{D}_i$}
            \STATE $\mathcal L_{local} = -\sum_{i \in \mathcal{D}_i}\log\frac{\sum_{\mathcal Z_L \in \mathcal{P}_j^L}\exp(sim(\xi_i, \mathcal Z_L))}{\sum_{\mathcal Z_L \in \mathcal{C}_j}\exp(sim(\xi_i, \mathcal Z_L))}$ by Eq.~\ref{loss_local_clusted_loss}
            \STATE $\mathcal L_{global} = -\sum_{i \in \mathcal{D}_i}\log\frac{\sum_{\mathcal Z_G \in \mathcal{P}_j^G}\exp(sim(\xi_i, \mathcal Z_G))}{\sum_{\mathcal Z_G \in \mathcal{U}_j}\exp(sim(\xi_i, \mathcal Z_G))}$ by Eq.~\ref{loss_global_clusted_loss}
            \STATE $\mathcal L = \frac {1}{|\mathcal{D}_i|} \sum_{i \in \mathcal{D}_i} F_i(\omega; \boldsymbol{x}_i)$ by Eq.~\ref{local_loss}
            \STATE $\mathcal L = \mathcal{L} + \mathcal{L}_{local} + \mathcal{L}_{global}$ by Eq.~\ref{obj:global_loss}
            \STATE {$\omega_i^t \gets \omega_i^t - \eta\nabla \mathcal L$}
        \ENDFOR
    \ENDFOR
    \STATE {Local clustering based on Eqs.~\ref{adjacency_matrix} and~\ref{average_first_partition}}
    \STATE {Obtain local signals $\bar{\xi}_{i}$ by Eq.~\ref{local_signal}}
    \RETURN $\omega_i^t$, $\bar{\xi}_{i}$
    \end{algorithmic}
\end{algorithm}

\subsection{Proposed Local Update Process}
\label{sub:proposed_local_updates}
Unlike prior research, which primarily emphasizes the global model, this paper shifts its focus to the performance of local models. Consequently, we introduce two extra loss functions for updating local models. First, we propose cross-client local clustered signals to enhance knowledge sharing among clients in Section~\ref{subsub:LCSC}. Second, we utilize global clustered signals to provide further guidance during the training of each local model in Section~\ref{subsub:GCSC}. Third, the overall objective for the local update is presented in Section~\ref{subsub:global_loss}.

\subsubsection{Local Clustered Signals Contrast}
\label{subsub:LCSC}
Within the FL environment, collaboration among local clients is a critical task because they collaborate to train models without sharing raw data. However, since the data distribution of each client is client-specific, the data between them may have different granularity. To leverage this granularity effectively, we propose a cross-client learning strategy based on contrasting local clustered signals. Moreover, by employing contrastive learning, we aim to ensure that the current representations are closer to local clustered signals belonging to the same class from all clients while being pushed further away from local clustered signals not belonging to that class from all clients. This approach can utilize data at a fine-grained level of granularity, helping to learn more discriminative representations and improving the model's ability to generalize across different clients.

Specifically, consider an instance ($\boldsymbol {x}_i, y_i) \in \mathcal D_{i}$, and then feed it into a neural network to obtain its feature vector $\xi_i = \boldsymbol f(\boldsymbol {x}_i)$ using Eq.~\ref{feature_extrac}. Subsequently, the feature vector is encouraged to be close to its respective local clustered signals with the same semantic class ($\mathcal{P}_j^L \in \xi_j$) while being discouraged from similarity with different semantics ($\mathcal{N}_j^L = \xi_j - \mathcal{P}_j^L$). Considering these factors, we introduce local clustered signals contrast, which can be defined as follows:
\begin{equation} \label{loss_local_clusted_loss}
    \mathcal L_{local} = -\sum_{i \in \mathcal{D}_i}\log\frac{\sum_{\mathcal Z_L \in \mathcal{P}_j^L}\exp(sim(\xi_i, \mathcal Z_L))}{\sum_{\mathcal Z_L \in \mathcal{C}_j}\exp(sim(\xi_i, \mathcal Z_L))},
\end{equation}
where $sim(\cdot)$ is the cosine similarity measurement between the query sample embedding $e_i$ with its corresponding local clustered signals $\mathcal Z_L \in \mathcal{P}_j^L$. Note that this optimization process is both communication-efficient and privacy-preserving, as the clustered local signals are only 2D data, and the cluster centers serve as pseudo-features, without information from the original data to some extent.

\subsubsection{Global Clustered Signals Contrast}
\label{subsub:GCSC}
Although the proposed local clustered signals contrast can facilitate cross-client knowledge sharing and enhance the discriminative abilities of local models, there is still a lack of a globally consistent signal to guide each local training. Therefore, to address this limitation, we further leverage the calculated global clustered signals from Eq.~\ref{avg_cluster_global_signals} to serve as a consistent signal to guide each local training in a contrastive learning manner. This approach can ensure intra-class compactness and inter-class separability in a global perspective, helping to mitigate the risk of overfitting to local optima and enhancing the model's generalizable ability.

Specifically, we introduce global clustered signals contrastive learning, where we pull the feature vector $e_i$ closer to its respective global clustered signals with the same semantics ($\mathcal{P}_j^G \in \mathcal{U}_j$) while pushing them away from those with different semantics ($\mathcal{N}_j^G = \mathcal{U}_j - \mathcal{P}_j^G$). Therefore, the global objective can be defined as follows:
\begin{equation} \label{loss_global_clusted_loss}
    \mathcal L_{global} = -\sum_{i \in \mathcal{D}_i}\log\frac{\sum_{\mathcal Z_G \in \mathcal{P}_j^G}\exp(sim(\xi_i, \mathcal Z_G))}{\sum_{\mathcal Z_G \in \mathcal{U}_j}\exp(sim(\xi_i, \mathcal Z_G))},
\end{equation}
where $sim(\cdot)$ is the cosine similarity measurement between the query sample embedding $e_i$ with corresponding global clustered signals $\mathcal Z_G \in \mathcal{P}_j^G$. Note that similar to the local clustered signals, the global clustered signals are also communication-efficient and privacy-preserving, as they undergo two rounds of clustering followed by an additional averaging process.

\subsubsection{Overall Objective Function}
\label{subsub:global_loss}
The proposed FedCCL framework consists of three components. The first component, denoted as $\mathcal L$, represents the loss associated with typical supervised learning tasks, and its calculation is detailed in Eq.~\ref{local_loss}. The second component, $\mathcal{L}_{local}$, is designed for local clustered signals contrastive learning, offering valuable cross-client insights and fine-grained intra-class information, as illustrated in Eq.~\ref{loss_local_clusted_loss}. The third component, $\mathcal{L}_{global}$, is utilized for global clustered signals contrastive learning, offering guidance for each local training, as illustrated in Eq.~\ref{loss_global_clusted_loss}. Therefore, every client can benefit from the three aforementioned loss functions, leading to the formulation of our overall objective function as follows:
\begin{equation} \label{obj:global_loss}
     \mathcal L = \mathcal L + \mathcal{L}_{local} + \mathcal{L}_{global}.
\end{equation}
An overall training process for FedCCL is shown in Algorithm~\ref{alg:FedCCL}. The input to FedCCL consists of heterogeneous datasets from multiple clients and their training parameters. The FedCCL algorithm comprises two main parts: a global server and local clients. The main steps of the global server part are executed from lines 1 to 9, while those of the local client part are executed from lines 1 to 13. In the local part, the local clustering and obtaining local signals are executed in lines 11 and 12, respectively. For each local training, the contrast between local clustered signals and global clustered signals is performed in lines 4 and 5, respectively, followed by the typical cross-entropy loss calculation in line 6. Finally, the overall objective for each client is summed in line 7, and then its model parameters are updated in line 8. In the server part, during each global communication round, each client conducts parallel local updates from lines 2 to 4. After this, the server aggregates all received local clustered signals in line 5. Following this, we perform global clustering in line 6 and obtain the global signals in line 7. Finally, we adopt a weighted average to incorporate local model updates into the global model and start the next communication round until convergence.

\begin{table*}
\centering
\caption{Top-1 average test accuracy (\%) of FedCCL and other baselines on Digit-5 and Office-10 under \textit{balanced inter-domain heterogeneity}. Avg represents the averaged value across all domains. The best results are in \textbf{bold}.}
\vspace{0.1cm}
\label{tab:balanced_inter_domain}
\resizebox{\linewidth}{!}{%
\begin{tabular}{r|ccccc|cc|cccc|cc} 
\toprule
& \multicolumn{7}{c|}{Digit-5} & \multicolumn{6}{c}{Office-10}\\
\cline{2-14}
\multirow{-2}{*}{Methods}  & MNIST  & SVHN  & USPS & Synth &  MNIST-M & Avg & $\triangle$ & Amazon  & Caltech  & DSLR  & Webcam & Avg & $\triangle$ \\
\hline
FedAvg  & 58.83  &  29.93 & 66.40  &  60.10 &  35.13 & 50.08 & - &  90.92  & 84.37  & 94.53 & 91.19  &  90.00  & - \\
FedProc & 63.27 & 29.00 & 68.53 & 63.47 & 35.80 & 52.01 & +1.93  & 90.88 & 84.23 & 93.13 & 89.61  &  89.71 & -0.29 \\
FedPCL  & 72.20 & 30.93 & 81.00 & 67.47 & 34.33 & 57.19 & +7.11 & 88.22 & 79.87 &  90.47  & 91.20 & 87.44  & -2.56  \\
FedProto &  60.27 & 29.67 & 66.33 & 60.80 & 30.60 &  49.53 & -0.55 &  94.08  & 86.42 & 97.28  & 95.44  & 93.05 & +3.05  \\
FPL & 71.27 & 27.53 & 81.47 & 67.53 & 37.33  & 57.03 & +6.95 & 87.42  & 81.08  & 90.29 & 87.01  & 86.70 & -3.30 \\
\hline
FedCCL & 75.93 & 29.47 & 84.67 & 67.40 & 35.27 & \textbf{58.55} & \textbf{+8.47} & 96.67 & 88.18  & 98.23  & 95.76  & \textbf{94.71} & \textbf{+4.71}  \\
\bottomrule
\end{tabular}}
\end{table*}

\begin{table*}
\centering
\caption{Top-1 average test accuracy (\%) of FedCCL and other baselines on Digit-5 and Office-10 under \textit{imbalanced inter-domain heterogeneity}. Avg represents the averaged value across all domains. The best results are in \textbf{bold}.}
\vspace{0.1cm}
\label{tab:imbalanced_inter_domain}
\resizebox{\linewidth}{!}{%
\begin{tabular}{r|ccccc|cc|cccc|cc} 
\toprule
& \multicolumn{7}{c|}{Digit-5} & \multicolumn{6}{c}{Office-10}\\
\cline{2-14}
\multirow{-2}{*}{Methods}  & MNIST  & SVHN  & USPS & Synth &  MNIST-M & Avg & $\triangle$ & Amazon  & Caltech  & DSLR  & Webcam & Avg & $\triangle$ \\
\hline
FedAvg &  29.67  & 14.97 &  38.04 & 24.83 & 15.33  & 24.57 & - & 76.55  & 74.88  & 85.68 & 82.49  &  79.90 & - \\
FedProc & 33.59  &  15.27 &  35.39 &  25.49  & 16.23 & 
25.39 & +0.82 &  77.80  & 73.65 & 84.18  & 78.36  &  78.50 & -1.40 \\
FedPCL & 49.85  & 14.92  &   48.87 &  28.13 & 17.66 &  
31.89 & +7.32 &  80.21  & 73.76 &  85.78 & 86.64 & 81.10  & +1.20 \\
FedProto  & 36.14  & 14.17  & 41.63  &  24.57  & 17.29  &  26.76 & +2.19 &  75.07 & 73.54  & 79.17  & 80.62  &  77.10 & -2.80 \\
FPL & 45.02  & 16.97 & 47.95 & 28.87 & 17.79  &  
31.72 & +7.15 & 82.91 &  79.58 & 86.19 & 82.97 & 83.16  & +3.26  \\
\hline
FedCCL & 48.15 & 16.08 & 51.33 & 37.30 & 18.27 & \textbf{34.03} & \textbf{+9.46} & 84.56 & 82.18 & 93.54  & 85.32  & \textbf{86.40} & \textbf{+6.50}   \\
\bottomrule
\end{tabular}}
\end{table*}

\begin{table*}[t]
\centering
\caption{Top-1 average test accuracy (\%) of FedCCL and other baselines on DomainNet under \textit{inter-domain heterogeneity} settings. Avg represents the averaged value across all domains. The best results are in \textbf{bold}.}
\vspace{0.1cm}
\label{tab:domainnet_combined}
\resizebox{\linewidth}{!}{%
\begin{tabular}{r|cccccc|c|cccccc|c}
\toprule
& \multicolumn{7}{c|}{Balanced} & \multicolumn{7}{c}{Imbalanced}  \\
\cline{2-15}
\multirow{-2}{*}{Methods} & Clipart & Infograph & Painting & Quickdraw & Real & Sketch & Avg & Clipart & Infograph & Painting & Quickdraw & Real & Sketch & Avg  \\
\hline
FedAvg & 70.80 & 39.20 & 76.60 & 50.60 & 81.40 & 68.00 & 64.43 & 42.80 & 30.30 & 42.10 & 22.30 & 46.80 & 41.80 & 37.68 \\
FedProc & 74.60 & 38.00 & 78.00 & 49.60 & 83.20 & 71.60 & 65.83 & 41.00 & 23.30 & 41.70 & 26.30 & 46.30 & 41.80 & 36.73 \\
FedPCL & 71.80 & 34.00 & 80.90 & 54.40 & 85.40 & 68.20 & 65.78 & 47.20 & 24.60 & 52.00 & 31.20 & 55.30 & 45.80 & 42.68 \\
FedProto & 74.40 & 35.00 & 75.60 & 42.80 & 84.80 & 65.40 & 63.00 & 53.00 & 28.80 & 55.70 & 27.90 & 62.20 & 49.70 & 46.21 \\
FPL & 74.60 & 34.60 & 84.20 & 56.40 & 88.80 & 71.60 & 68.36 & 49.40 & 30.60 & 52.30 & 31.80 & 61.70 & 47.30 & 45.51 \\
\hline
FedCCL & 77.20 & 37.80 & 84.80 & 62.00 & 89.80 & 73.60 & \textbf{70.86} & 56.60 & 31.30 & 61.50 & 25.20 & 69.00 & 56.10 & \textbf{49.95} \\
\bottomrule
\end{tabular}}
\end{table*}

\section{Experiments}
\label{sec:experiments}
In this section, we will first outline the experimental setup in Section~\ref{subsec:experimental_setup}. Following that, we proceed with the performance comparison in Section~\ref{subsec:accuracy_compar}, followed by the communication efficiency comparison in Section~\ref{subsec:communi_effici}. Subsequently, an ablation study is performed in Section~\ref{subsec:ablation_study} to analyze different configurations of our algorithm and their impact on performance. Finally, we conduct the robustness evaluation in Section~\ref{subsec:robust_compar}.

\subsection{Experimental Setup}
\label{subsec:experimental_setup}
\textbf{Datasets.} We conduct experiments on six popular benchmark datasets: MNIST~\cite{lecun1998gradient}, Fashion-MNIST~\cite{xiao2017fashion}, CIFAR-10~\cite{krizhevsky2009learning}, Digit-5~\cite{zhou2020learning}, Office-10~\cite{gong2012geodesic}, and DomainNet~\cite{peng2018synthetic} to verify the potential benefits of FedCCL. The first three datasets are utilized to evaluate performance under intra-domain heterogeneity, given that they exclusively contain samples from the same domain. Conversely, the last three datasets, each comprising data from different domains, are employed to evaluate performance under inter-domain heterogeneity. Specifically, \textbf{MNIST}~\cite{lecun1998gradient} is a dataset of handwritten digits containing 10 classes, while \textbf{Fashion-MNIST}~\cite{xiao2017fashion} contains a total of 10 classes, including different types of clothing items. \textbf{CIFAR-10}~\cite{krizhevsky2009learning} is a more complex dataset, where each sample is a 32x32x3 color image, consisting of 10 classes. \textbf{Digit-5}~\cite{zhou2020learning} is a comprehensive dataset that comprises images of handwritten digits gathered from five widely recognized datasets, including SVHN, USPS, MNIST, MNIST-M, and Synth. \textbf{Office-10}~\cite{gong2012geodesic} comprises images from four distinct office environments: Amazon, Webcam, DSLR, and Caltech, for a total of 10 categories, derived from the overlap between Office31~\cite{saenko2010adapting} and Caltech-256~\cite{griffin2007caltech} datasets. \textbf{DomainNet} is a large-scale image classification dataset with over 600,000 images spanning 345 categories. It comprises six domains: clipart, infograph, painting, quickdraw, real, and sketch.

\begin{table}[t]
\centering
\caption{Top-1 average test accuracy (\%) of FedCCL and other baselines on MNIST, Fashion-MNIST, and CIFAR-10 under \textit{imbalanced intra-domain heterogeneity}. Avg represents the averaged value across all domains. The best results are in \textbf{bold}.}
\label{tab:intra_domain_m_fm_cifar_comparison}
\vspace{0.1cm}
\resizebox{0.6\linewidth}{!}{%
\begin{tabular}{r|cc|cc|cc} 
\toprule
Methods & MNIST & $\triangle$ & FMNIST& $\triangle$ & CIFAR-10&$\triangle$\\ 
\midrule
FedAvg & 94.45 & - & 73.66 & - & 57.65  & - \\
FedProc & 94.75  & +0.30 & 77.65 & +3.99 & 54.08 & -3.57 \\
FedPCL & 94.89 & +0.44 & 72.33 & -1.33 & 60.16 & +2.51 \\
FedProto & 94.30 & -0.15 & 72.86 & -0.80 & 60.72 & +3.07 \\
FPL  & 93.47 & -0.98  & 74.52 & +0.86 & 59.39 & +1.74\\
\hline
FedCCL & \textbf{95.15} & \textbf{+0.70} & \textbf{80.02}  & \textbf{+6.36} & \textbf{61.90}  &  \textbf{+4.25} \\
\bottomrule
\end{tabular}}
\end{table}

\textbf{Baselines and Implement Details.} We compare our proposal with several popular baselines, including FedAvg~\cite{mcmahan2017communication},  FedPCL~\cite{tan2022federated} with model parameter averaging, FedProc~\cite{mu2023fedproc}, FedProto~\cite{tan2022fedproto} with model parameter averaging, and FPL~\cite{huang2023rethinking}. To evaluate the performance of our proposal within the imbalanced intra-domain heterogeneity setting, we focus on MNIST, Fashion-MNIST, and CIFAR-10 datasets, as they all belong to the same domain. By default, we opt for 10 clients for these experiments. For experiments conducted within the intra-domain setting, our focus is the Digit-5, Office-10, and DomainNet datasets, as each dataset contains varying numbers of sub-datasets from different domains. Therefore, we utilize 5, 4, and 6 clients to simulate balanced and unbalanced inter-domain heterogeneity scenarios for Digit-5, Office-10, and DomainNet, respectively. Besides, we use Dirichlet~\cite{yurochkin2019bayesian} distribution to stimulate the imbalanced setting, where the magnitude of the Dirichlet parameter value is positively related to the degree of imbalance. Here, we set the default value to 0.5 for the imbalanced setting.

Moreover, similar to prior works such as~\cite{li2021model,li2020federated}, we employ a 4-layer CNN model for MNIST and Fashion-MNIST, and a 5-layer CNN model for CIFAR-10, with decision-making layers of both CNN models having a dimension of 128. For inter-domain datasets, following approaches like~\cite{tan2022federated,liu2020universal}, we utilize a pre-trained Tiny-ViT~\cite{wu2022tinyvit} as the feature extractor layers, followed by a two-layer MLP with an input dimension of 192 and a hidden dimension of 128 for the decision-making layers. Note that all methods employ the same network architecture to ensure a fair comparison across various tasks. Given that the CNN model for intra-domain datasets requires training from scratch, we set the default global communication round to 100, and the number of local epochs is configured as 1 for MNIST and Fashion-MNIST, and 5 for CIFAR-10. In addition, for inter-domain datasets, as we opt for a pre-trained model to enhance training efficiency, we set the default global communication round to 60, with the number of local epochs configured as 1. We utilize the SGD optimizer with a learning rate of 0.01 for the CNN model, the Adam optimizer with a learning rate of 0.001 for the pre-trained model, and the batch size for all sets to 64. By default, we set the temperature to 0.07 for all experiments like~\cite{he2020momentum,wu2018unsupervised}. Experimental results are reported based on three independent experiments, using the last five communication rounds as the final performance.

\subsection{Performance Comparison}
\label{subsec:accuracy_compar}
We conduct a thorough evaluation across six tasks to compare our approach with several baselines under both intra- and inter-domain heterogeneity. For inter-domain tasks, the results are presented in Table~\ref{tab:balanced_inter_domain}, Table~\ref{tab:imbalanced_inter_domain}, and Table~\ref{tab:domainnet_combined}. In the case of intra-domain tasks, the results are reported in Table~\ref{tab:intra_domain_m_fm_cifar_comparison}. For each task, we not only present the accuracy under balanced settings but also under imbalanced settings. An overall observation reveals that our proposal consistently outperforms all baselines across various tasks.

Specifically, considering the inter-domain scenarios, we take the results of Digit-5 in Table~\ref{tab:balanced_inter_domain} and Table~\ref{tab:imbalanced_inter_domain} as an example, it is evident that FedCCL outperforms all the baselines by a significant margin under both balanced and imbalanced settings. For example, in the balanced setting, FedCCL shows a remarkable 8.47\% improvement over the widely-used baseline FedAvg, while still surpassing the second-best FedPCL by 1.36\%. Additionally, under the imbalanced setting, FedCCL demonstrates a 9.46\% increase over FedAvg and still has a 2.14\% improvement compared to the second-best FedPCL. Moreover, it can also be observed that all methods experience a drop in accuracy, including ours, in imbalanced settings. This further confirms that imbalanced settings pose more challenges than balanced settings. Notably, despite this challenge, our proposal still outperforms others even in an imbalanced settings. Similar results are also evident in Office-10 and DomainNet, which highlights the effectiveness of our proposal in handling both intra-domain and inter-domain heterogeneity. 

To obtain a comprehensive evaluation under domain heterogeneity, we also present the performance based on three popular datasets: MNIST, Fashion-MNIST, and CIFAR-10, as shown in Table~\ref{tab:intra_domain_m_fm_cifar_comparison}. It is worthwhile noting that since MNIST and Fashion-MNIST are considered relatively easier datasets compared to CIFAR-10, we set the Dirichlet parameter, which controls the degree of imbalance, to a value 10 times smaller, specifically 0.05. This adjustment introduces more challenges, as a smaller Dirichlet parameter value indicates a more imbalanced data distribution. Specifically, our proposal demonstrates a 6.36\% higher accuracy compared to FedAvg, and still maintains a 2.37\% lead over the second-best approach, FedProc. Similar trends are observed in MNIST and CIFAR-10 datasets. Although the improvement in MNIST is not substantial, this could be attributed to the inherent simplicity of the MNIST dataset, making it difficult to discern significant differences among approaches. In general, the above results and findings further support our motivation that local clustered signals provide rich knowledge sharing among clients while global clustered signals offer a consistent optimization direction. Both of these strategies contribute to the effectiveness of our approach in improving performance across various datasets.

\begin{figure}[t]
\centering
\includegraphics[width=0.50\textwidth]{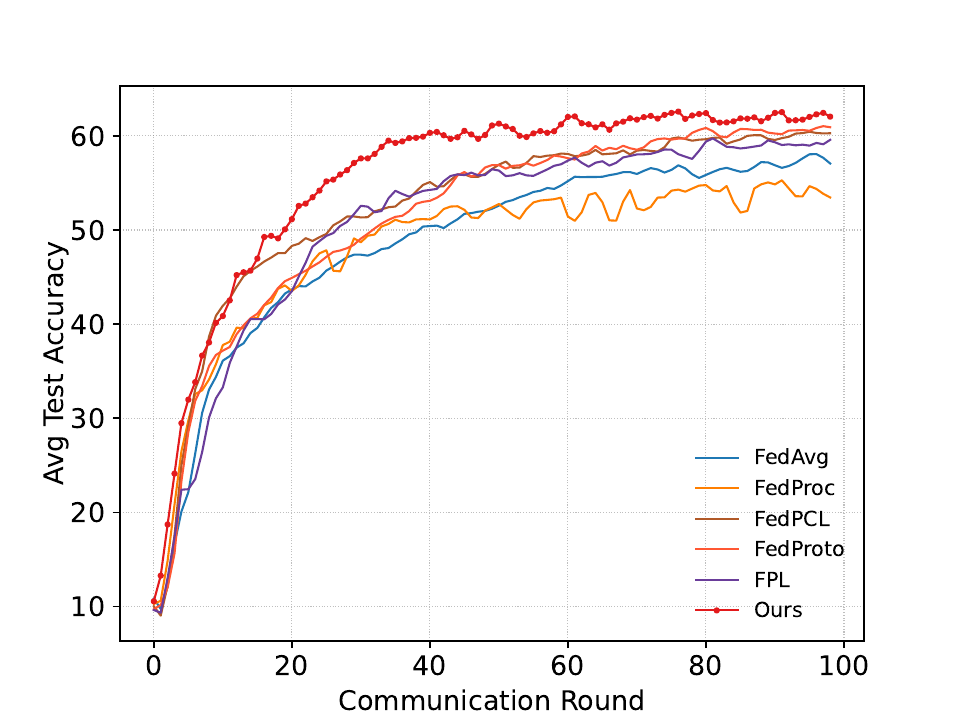}
\caption{Comparison of communication efficiency (\%) on CIFAR-10 task under \textit{imbalanced intra-domain heterogeneity} scenarios.}
\label{fig:com_effici_cifar10_nonIID}
\end{figure}

\begin{figure}[t]
\centering
\includegraphics[width=0.50\textwidth]{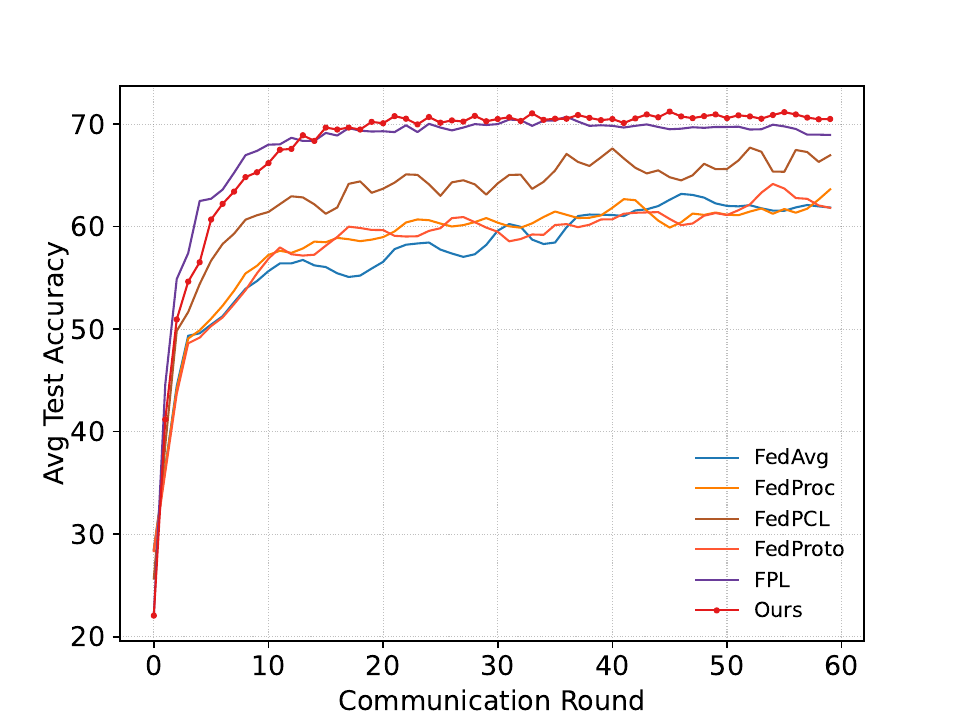}
\caption{Comparison of communication efficiency (\%) on DomainNet task under \textit{balanced inter-domain heterogeneity} scenarios.}
\label{fig:com_effici_domainnet_nonIID}
\end{figure}

\begin{figure}[t]
\centering
\includegraphics[width=0.50\textwidth]{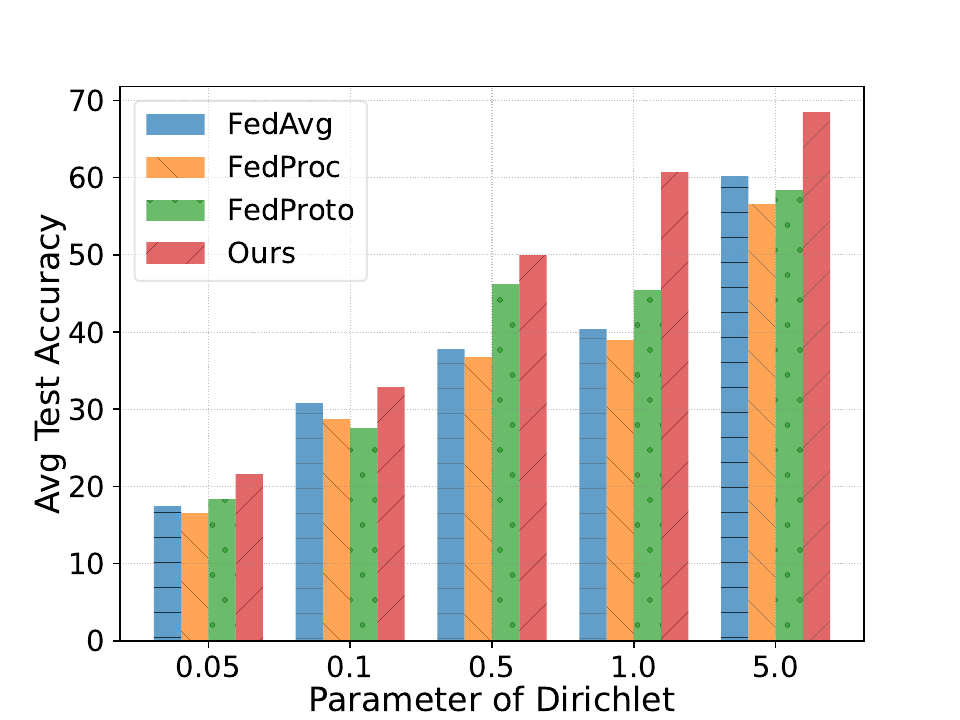}
\caption{Comparison of average test accuracy (\%) on DomainNet under \textit{imbalanced inter-domain heterogeneity} scenarios. A lower Dirichlet parameter value indicates server heterogeneity.}
\label{fig:diff_dirichlet_domainnet_nonIID}
\end{figure}

\begin{figure}[t]
\centering
\includegraphics[width=0.50\textwidth]{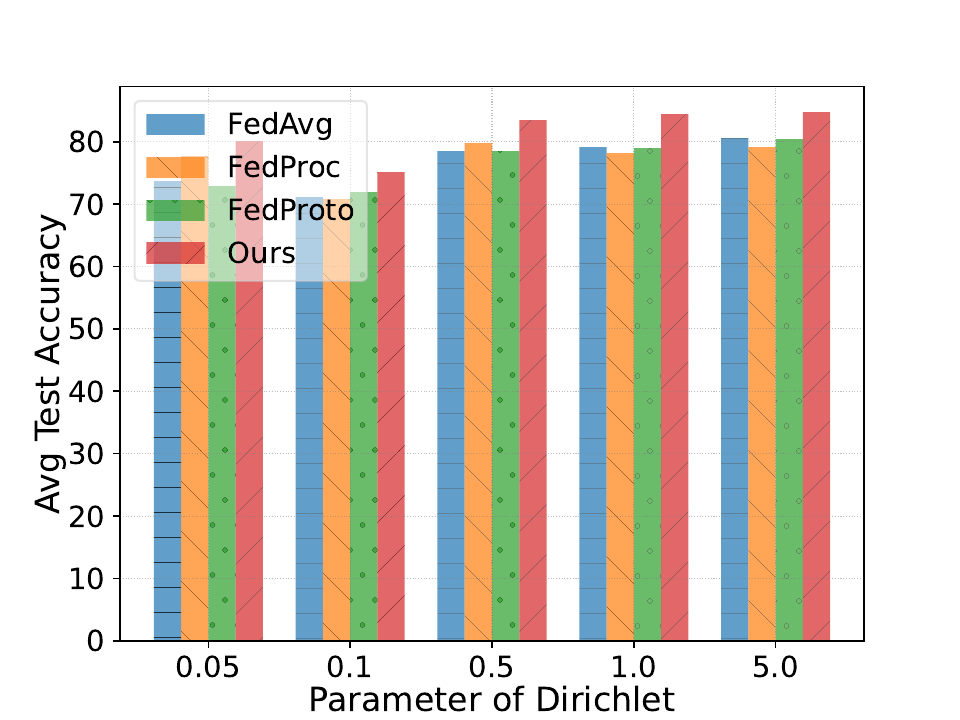}
\caption{Comparison of average test accuracy (\%) on Fashion-MNIST under \textit{imbalanced intra-domain heterogeneity} scenarios. A lower Dirichlet parameter value indicates server heterogeneity.}
\label{fig:diff_dirichlet_fashionmnist_nonIID}
\end{figure}

\begin{table}[t]
\centering
\caption{Comparison of different configurations on Digit-5 and Office-10 datasets under \textit{balanced and imbalanced inter-domain heterogeneity}. "Local" means only local signals contrast is used for local training, "Global" means only global signals contrast is used for local training.}
\vspace{0.1cm}
\label{tab:ablation_digit_office_comparison}
\resizebox{0.60\linewidth}{!}{
\begin{tabular}{c|cc|cc|cc}
\toprule
Dataset & Local & Global & Balanced  & $\triangle$ & Imbalanced  & $\triangle$ \\
\midrule
Digit-5 & $\times$ & $\times$ & 50.08 & - & 24.57 & - \\
& \checkmark & $\times$ & 56.01 & +5.93 & 30.12 & +5.55 \\
& $\times$ & \checkmark & 57.94 & +7.86 & 31.90 & +7.33 \\
& \checkmark & \checkmark & \textbf{58.55} & \textbf{+8.47} & \textbf{34.03} & \textbf{+9.46}\\
\midrule
Office-10 & $\times$ & $\times$ & 90.00 & - & 79.90 & -\\
& \checkmark & $\times$ & 90.61 & +0.61 & 83.65& +3.75 \\
& $\times$ & \checkmark & 93.40 & +3.40 & 85.14 & +5.24 \\
& \checkmark & \checkmark & \textbf{94.71}& \textbf{+4.71} & \textbf{86.40}& \textbf{+6.50} \\
\bottomrule
\end{tabular}}
\end{table}

\begin{table}[h]
\centering
\caption{Comparison of different configurations on Fashion-MNIST and CIFAR-10 datasets under \textit{intra-domain heterogeneity}. "Local" and "Global" have the same meaning as Table~\ref{tab:ablation_digit_office_comparison}.}
\vspace{0.1cm}
\label{tab:ablation_mnist_cifar10_comparison}
\resizebox{0.60\linewidth}{!}{
\begin{tabular}{cc|cc|cc}
\toprule
Local & Global & Fashion-MNIST  & $\triangle$ & CIFAR-10  & $\triangle$ \\
\midrule
$\times$ & $\times$ & 78.38 & - & 57.65 & - \\
\checkmark & $\times$ & 82.64  & +4.26  & 60.16 & +2.51 \\
$\times$ & \checkmark & 82.45  & +4.07  & 61.29 & +3.64 \\
\checkmark & \checkmark & \textbf{83.41} & \textbf{+5.03} & \textbf{61.90} & \textbf{+4.25}\\
\bottomrule
\end{tabular}}
\end{table}

\begin{table}[t]
\centering
\caption{Comparison of different contrast strategies on the Digit-5 and Office-10 datasets. "Local" indicates the usage of local signals average or local signals clustered contrast for local training under imbalanced inter-domain heterogeneity conditions, while "Global" denotes the utilization of global signals average or global signals clustered contrast for local training under balanced inter-domain heterogeneity.}
\vspace{0.1cm}
\label{tab:avg_or_cluster_digit_office_comparison}
\resizebox{0.60\linewidth}{!}{
\begin{tabular}{c|cc|cc|cc}
\toprule
Contrast & Avg & Cluster & Digit-5 & $\triangle$ & Office-10 & $\triangle$ \\
\midrule
Local & \checkmark & & 27.37  & - & 82.50  &  -\\
& & \checkmark & \textbf{34.03} & \textbf{+6.66} & \textbf{86.40}  & \textbf{+3.90} \\
\midrule
Global & \checkmark &  & 57.57  &  - & 91.90  &  - \\
&  & \checkmark & \textbf{58.55} & \textbf{+0.98} & \textbf{94.71} & \textbf{+2.81} \\
\bottomrule
\end{tabular}}
\end{table}

\begin{table}[t]
\centering
\caption{Robustness comparison (\%) for various tasks across different metrics. ``FGSM" is the fast gradient sign method (FGSM)~\cite{goodfellow2014explaining}, and ``PGD-20'' is the projected gradient descent (PGD)~\cite{madry2017towards} algorithm with 20 steps.}
\vspace{0.1cm}
\label{tab:robustness_comparison}
\resizebox{0.56\linewidth}{!}{
\begin{tabular}{l|l|ccc}
\toprule
Dataset & Method & Clean & FGSM & PGD-20 \\ 
\midrule
& FedAvg & 95.78 & 42.64 & 0.37 \\
MNIST & FedAvg+ & 94.43 & 58.98 & 0.62  \\
& FedCCL+ & \textbf{96.66}  & \textbf{69.90}  & \textbf{0.77}  \\ 
\midrule
& FedAvg & 78.77 & 36.16 & 0.17 \\
Fashion-MNIST & FedAvg+ & 62.32 & 47.65 & 0.43  \\
& FedCCL+ & \textbf{67.64}  & \textbf{54.97}  & \textbf{0.51}  \\ 
\bottomrule
\end{tabular}}
\end{table}

\subsection{Communication Efficiency}
\label{subsec:communi_effici}
To illustrate the communication efficiency of our proposal and several baselines, we present the results of CIFAR-10 and DomainNet in Figure~\ref{fig:com_effici_cifar10_nonIID} and Figure~\ref{fig:com_effici_domainnet_nonIID}, respectively, for each communication round. From the results shown in the figure, an overall trend demonstrates that our proposal can achieve higher test accuracy and faster convergence compared to several baselines.

Specifically, in Figure~\ref{fig:com_effici_cifar10_nonIID}, our proposal significantly outperforms other baselines starting from approximately the 10th communication round for CIFAR-10 tasks. Additionally, our approach achieves a notable improvement of approximately 2\% compared to the second-best performing method, FedProto, after 100 communication rounds. For the DomainNet task in Figure~\ref{fig:com_effici_domainnet_nonIID}, our method demonstrates a relatively fast convergence rate and accuracy, starting from the 15th round of communication, compared to other baseline methods except for the second-best performing method, FPL. However, it is worthwhile to note that although FPL performs comparably with us in the first 40 rounds of communication, our method still achieves approximately a 2\% improvement over FPL after 60 rounds of communication. Furthermore, we highlight that our proposal experiences a relatively lower variance compared to all the baselines, to some extent.

\subsection{Ablation Study}
\label{subsec:ablation_study}
\textbf{Effects of Key Components.} To analyze the effectiveness of components within our proposed federated framework, we conduct an ablation study to evaluate the impact of each component on the overall performance. Table~\ref{tab:ablation_digit_office_comparison} and Table~\ref{tab:ablation_mnist_cifar10_comparison} present the results of the ablation analysis on inter- and intra-domain tasks, respectively. From the results in the table, we have several observations. First, when local signals and global signals are not utilized for local training, the performance across various tasks is degraded. This confirms the essential role of both local and global signals in the training process. Second, either local signals or global signals can significantly improve the performance. This shows that local training can benefit from local signals and global signals. Third, better performance can be obtained by combining local and global signals, which supports our motivation that local signals can capture rich knowledge while global signals can be used to guide each local training. Taking the results of Digit-5 from Table~\ref{tab:ablation_digit_office_comparison} as an example, without local and global signals, the performance in both balanced and imbalanced settings is notably lower, at 50.08 and 24.57, respectively. However, incorporating either local or global signals leads to an improvement of at least 5\%. Moreover, by combining both local and global strategies, FedCCL achieves a performance gain of 8.47\% and 9.46\% in balanced and imbalanced settings, respectively.

\textbf{Comparison to Averaged Signals.} To validate the effectiveness of the adopted cluster signals, we compare local averaged signals and local cluster signals, as well as global averaged signals and global cluster signals in Table~\ref{tab:avg_or_cluster_digit_office_comparison}. From the results, several observations can be made. First, both local clustering and global clustering demonstrate benefits to performance, confirming the significance of clustering in local training. Second, replacing either local or global signals with the averaged signal results in a decrease in accuracy. This highlights the advantageous role of local and global clustering in assisting local training. For example, in the Office-10 task, the local cluster demonstrates a significant 3.90\% performance improvement compared to the local averaged strategy, and the global cluster shows a 2.81\% improvement over the global averaged strategy. Overall, the analysis of different components as well as the comparison to the averaged signal validates, to some extent, our motivation. It suggests that local signals can promote knowledge sharing among clients, while global signals effectively guide each local training process. Therefore, both local signals and global signals contribute to the overall performance improvement in federated environments.

\subsection{Robustness Evaluation}
\label{subsec:robust_compar}
\textbf{Robust to Different Heterogeneity.} As highlighted in the previous section, imbalanced settings pose significant challenges in federated environments. Given the varying degrees of data heterogeneity in real-world scenarios, it becomes crucial to evaluate the robustness of an algorithm across these conditions for real-world deployments. We employ the Dirichlet distribution to simulate different degrees of heterogeneity in the real world. The results for different tasks are shown in Figure~\ref{fig:diff_dirichlet_domainnet_nonIID} and Figure~\ref{fig:diff_dirichlet_fashionmnist_nonIID}. From the results, several observations can be drawn. First, with a smaller Dirichlet value, all methods, including ours, experience a decline in accuracy in most cases, thereby confirming the challenges posed by domain heterogeneity. Second, our proposed method outperforms several baselines in both intra-domain and inter-domain settings. For example, in the DomainNet task, when the parameter is set to 0.05 compared to when it is set to 5.0, the former exhibits a degree of heterogeneity that is 100x greater than the latter. Consequently, the accuracy of FedCCL decreases from 68.40 to 21.55. However, even in such extreme cases, our proposal outperforms the second-best method, FedAvg, by 8.2\% with a parameter of 5.0, and surpasses the second-best method, FedProto, by 3.29\% with a parameter of 0.05. This underscores the effectiveness of our approach, which exhibits robustness to varying degrees of heterogeneity compared to other methods.

\textbf{Robust to Adversarial Attacks.} To evaluate the robustness of our proposal against adversarial attacks, we first provide the common practice (i.e., adversarial training (AT)) against adversarial attacks in~\ref{appendix:at}. Subsequently, we provide an enhanced version of our proposal, FedCCL+, that incorporates AT to defend against adversarial attacks in~\ref{appendix:robust_ours_plus}. The comparison of robustness between our proposal and the baseline, FedAvg+, is depicted in Table~\ref{tab:robustness_comparison}. Note that FedAvg+ is an enhanced version that incorporates the AT strategy into the vanilla FedAvg framework. In addition, the metrics ``clean", ``FGSM", and ``PGD-20" represent the model after the AT process is tested on clean images and adversarial images that are perturbed by FGSM and PGD-20 algorithms, respectively. From the results in the table, several observations can made. First, the accuracy of all methods, including ours, decreases when subjected to adversarial attacks. For example, in the MNIST task, FedCCL+ dropped from 96.66\% to 69.90\% and 0.77\% after FGSM and PGD-20 attacks, respectively. Similarly, FedAvg+ experienced a decrease from 94.43\% to 58.98\% and 0.62\% after the same attacks, respectively. Second, FedCCL+ exhibits relatively higher robustness than other baselines across various tasks. For example, in the Fashion-MNIST task, FedCCL+ outperforms FedAvg+ by approximately 7\% under the FGSM attack.

\section{Conclusion}
\label{sec:conclusion}
In this paper, we first outline the non-IID challenges from an intra-domain and inter-domain perspective. Subsequently, with the goal of learning a well-generalizable model for each client, we introduce a dual-clustered signal contrast strategy, which consists of two key components. First, to facilitate knowledge sharing among clients, we propose cross-client local clustered signals contrastive learning, where local signals serve as a carrier for fine-grained intra-class knowledge sharing and facilitate cross-client knowledge exchange. Second, to provide guidance for each local training process, we introduce global signals. These signals provide a consistent optimization direction for each client's training, thereby enhancing its generalizable abilities. Experimental results on various tasks from different domains heterogeneity demonstrate that our approach achieves comparable or superior performance compared to several baselines. Moreover, we also provide preliminary results demonstrating the robustness of our strategy against adversarial attacks after AT.

\section*{Acknowledgments}
This work was supported by ....

\bibliographystyle{unsrt}  
\bibliography{bib_local,bib_global} 
\appendix
\section{Robustness to Adversarial Attacks}
In this section, we first outline the main process for adversarial training in~\ref{appendix:at}. Subsequently, we present the enhanced version of our proposal in~\ref{appendix:robust_ours_plus}.

\subsection{Adversarial Training}
\label{appendix:at}
Adversarial training~\cite{athalye2018robustness} is an effective strategy for defending against various adversarial attacks. This strategy is typically formulated as a min-max problem, aiming to minimize prediction errors against adversaries while maximizing adversarial loss by interfering with the input. This process can be formulated as follows:
\begin{equation} \label{AT_federated}
   \mathop {\min}_{\omega} \mathbb E_{(\boldsymbol{x}, y) \sim \mathcal{D}} \left[\mathop {\max}_{||\delta||_\infty \leq \epsilon} \mathcal{L}(\omega; \boldsymbol{x} + \delta, y)\right],
\end{equation}
where $\delta$ represents the perturbation that is imperceptible (or quasi-imperceptible) to the human eye, and is constrained by an upper bound $\epsilon$ on the $\ell_\infty$-norm, i.e., $||\delta||_\infty \leq \epsilon$.

The PGD algorithm is a common approach used to address the inner maximization component, while the outer part is handled by gradient descent such as SGD or Adam, which is a standard model training process. Therefore, we illustrate the inner part optimization process as follows:
\begin{equation}
\label{eq:pgd}
\boldsymbol x^{t+1}=\Pi_{\boldsymbol{x} + \delta}\left(\boldsymbol x^t+\alpha\sign(\nabla_{\boldsymbol x^t}  \mathcal{L}(\omega; \boldsymbol{x}^t, y) \right),
\end{equation}
where $\sign(\cdot$) is the indicator function, $\Pi$ denotes the projection function, and $\boldsymbol x_i^t$ is the generation for adversarial examples generation at step $t$. Note that adversarial examples are typically generated by adding perturbations $\delta$ to clean examples $\boldsymbol{x}$.

\subsection{Robust FedLCC+ Framework}
\label{appendix:robust_ours_plus}
To design a robust federated framework, one effective approach is to incorporate adversarial training into the local model training process. Therefore, a common strategy involves directly integrating adversarial training into the vanilla FL framework, as follows:
\begin{equation} \label{adv_local_loss}
   {\hat{\mathcal{L}_{ce}}} = \frac {1}{|\mathcal{D}|} \sum_{i \in \mathcal{D}} F(\omega; \widetilde{\boldsymbol{x}_i}),
\end{equation}
where $\widetilde{\boldsymbol{x}}$ is the generated adversarial examples by PGD algorithm. Note that the objective of Eq.~\ref{adv_local_loss} is to optimize the model such that it can still make accurate predictions even under perturbed inputs $\widetilde{\boldsymbol{x}}$. By achieving this, the model would exhibit robustness against adversarial attacks.

However, in the FL environment, where data distribution bias commonly exists, we argue that directly training the model using Eq.~\ref{adv_local_loss} would encounter significant challenges in achieving an unbiased robust model due to the non-IID data distribution. Therefore, we preliminarily propose an enhanced version of FedCCL that could potentially demonstrate robustness against attacks, as follows.

Inspired by logit pairing proposed by~\cite{kannan2018adversarial}, we can encourage the adversarial feature by the feature extractor layers to be close to local/global signals obtained on the corresponding original images. Then, the local cluster contrast and global cluster contrast loss function can be rewritten as follows:
\begin{equation} \label{adv_loss_local_clusted_loss}
    \hat{\mathcal L_{local}} = -\sum_{i \in \mathcal{D}_i}\log\frac{\sum_{\mathcal Z_L \in \mathcal{P}_j^L}\exp(sim(\hat{\xi_i}, \mathcal Z_L))}{\sum_{\mathcal Z_L \in \mathcal{C}_j}\exp(sim(\hat{\xi_i}, \mathcal Z_L))},
\end{equation}
\begin{equation} \label{adv_loss_global_clusted_loss}
    \hat{\mathcal L_{global}} = -\sum_{i \in \mathcal{D}_i}\log\frac{\sum_{\mathcal Z_G \in \mathcal{P}_j^G}\exp(sim(\hat{\xi_i}, \mathcal Z_G))}{\sum_{\mathcal Z_G \in \mathcal{U}_j}\exp(sim(\hat{\xi_i}, \mathcal Z_G))},
\end{equation}
where $\hat{\xi_i}$ denotes the adversarial feature generated by the feature extractor layers. Equations~\ref{adv_loss_local_clusted_loss} and~\ref{adv_loss_global_clusted_loss} represent the redesigned loss functions, aimed at acquiring a robust and well-generalizable local model.

Finally, the overall loss function for training a robust and well-generalizable local model is shown in Eq.~\ref{obj:global_adv_loss}. 
\begin{equation} \label{obj:global_adv_loss}
     \hat{\mathcal L} = {\hat{\mathcal{L}_{ce}}} + \hat{\mathcal{L}_{local}} + \hat{\mathcal{L}_{global}}.
\end{equation}
Based on this formulation, we require the classifier not only to classify adversarial examples correctly but also to maintain consistency between the adversarial signals and those obtained from the corresponding original images. In other words, the classifier's classification results for the same image should be consistent on both the clean image and the adversarial example. 

\end{document}